\documentclass[11pt,a4paper]{article}
\usepackage[hyperref]{acl2020}
\usepackage{times}
\usepackage{latexsym}
\usepackage{url}

\usepackage{multirow}
\usepackage{booktabs}
\usepackage{wrapfig}
\usepackage{multicol}
\usepackage{float}
\usepackage{graphicx}
\usepackage{amsmath}
\usepackage{amsfonts}
\usepackage{amssymb}
\usepackage{amsthm}
\usepackage{float}
\usepackage{breqn}
\usepackage{todonotes}
\usepackage{subcaption}
\usepackage{adjustbox}
\usepackage{multicol}
\usepackage{bbm}
\usepackage{enumitem}

\usepackage{microtype}
\usepackage{soul}

\aclfinalcopy 

\title{Humor@IITK at SemEval-2021 Task 7: Large Language Models for Quantifying Humor and Offensiveness}

\author{Aishwarya Gupta$^{*}$,\ \ \
  Avik Pal$^{*}$,\ \ \     
  Bholeshwar Khurana$^{*}$,\ \ \    
  Lakshay Tyagi\thanks{\quad Authors contributed equally to the work. Names is alphabetical order.},\\ 
  \large{\textbf{Ashutosh Modi}} \\
{Indian Institute of Technology Kanpur (IIT Kanpur)} \\
  {\tt \{aishwaryag20, avikpal, bholek, lakshayt\}@iitk.ac.in}  \\
  {\tt ashutoshm@cse.iitk.ac.in}  \\
} 

\date{}

\begin{document}
\maketitle
\begin{abstract}
Humor and Offense are highly subjective due to multiple word senses, cultural knowledge, and pragmatic competence. Hence, accurately detecting humorous and offensive texts has several compelling use cases in Recommendation Systems and Personalized Content Moderation. However, due to the lack of an extensive labeled dataset, most prior works in this domain haven't explored large neural models for subjective humor understanding. This paper explores whether large neural models and their ensembles can capture the intricacies associated with humor/offense detection and rating. Our experiments on the SemEval-2021 Task~7: HaHackathon show that we can develop reasonable humor and offense detection systems with such models. Our models are ranked third in subtask 1b and consistently ranked around the top $33\%$ of the leaderboard for the remaining subtasks. 
\end{abstract}

\section{Introduction}

Like most figurative languages, humor/offense pose interesting linguistic challenges to Natural Language Processing due to its emphasis on multiple word senses, cultural knowledge, sarcasm, and pragmatic competence.  A joke's perception is highly subjective, and age, gender, and socioeconomic status extensively influence it. Prior humor detection/rating challenges treated humor as an objective concept. SemEval 2021 Task 7 \cite{semeval2021task7} is the first humor detection challenge that incorporates the subjectivity associated with humor and offense across different demographic groups. Users from varied age groups and genders annotated the data with the text's humor and have provided an associated score for the same. It is also quite a generic phenomenon that a text might be humorous to one and normal/offensive to another. Rarely has it been noticed that the same content is globally accepted as witty. To the best of our knowledge, \citet{semeval2021task7} is the first initiative towards annotating the underlying humor as controversial or not. Understanding whether a text is humorous and/or offensive will aid downstream tasks, such as personalized content moderation, recommendation systems, and flagging offensive content.

Large Language Models (LLMs) have recently emerged as the SOTA for various Natural Language Understanding Tasks~\citep{lewis2019bart, raffel2019exploring, conneau2019unsupervised, zhang2020pegasus}. However, typical day-to-day texts, where these models have shown state of the art performance, are less ambiguous than texts having puns/jokes. Training and evaluating LLMs in the context of highly ambiguous/subjective English texts would serve as an excellent benchmark to figure out the current shortcomings of these models.  This paper studies various large language models -- BERT \citep{devlin2018bert}, RoBERTa \citep{liu2019roberta}, XLNet \citep{yang2019xlnet}, ERNIE-2.0 \citep{sun2019ernie} and DeBERTa \citep{he2020deberta} and their ensembles -- for humor and offense detection tasks. Additionally, we explore a Multi-Task Learning framework to train on all the four sub-tasks jointly and observe that joint training improves the performance in regression tasks.

We have achieved significant performance on all the subtasks and have consistently ranked $\sim \frac{1}{3}^{rd}$ of the total submissions. We were ranked (1) $21^{st}$ with an F-score and accuracy of $94.8\%$ and $95.81\%$ respectively in Task 1a, (2) $3^{rd}$ with an RMSE score of $0.521$ in Task 1b, (3) $9^{th}$ with an F-score and accuracy of $45.2\%$ and $62.09\%$ respectively in Task 1c; and (4) $16^{th}$ with an RMSE score of $0.4607$ in Task 2. We release the code for models and experiments via GitHub\footnote{ \url{https://github.com/aishgupta/Quantifying-Humor-Offensiveness}}

We organize the rest of the paper as: we begin with a description of the challenge tasks followed by a brief literature survey in section~\ref{sec:bg}. We then describe all of our proposed models in section~\ref{sec:system} with training details in section~\ref{sec:expt} and present the experimental results in section~\ref{sec:results}. Finally, we analyze our findings and conclude in section~\ref{sec:analysis}, and \ref{sec:conclude} respectively.

\section{Background}
\label{sec:bg}

\subsection{Problem Description}

SemEval 2021 Task 7: HaHackathon: Detecting and Rating Humor and Offense~\citep{semeval2021task7} involves two main tasks -- humor detection and offense detection. The organizers further subdivide the task into following subtasks:

\begin{enumerate}
\item Humor detection tasks:
\begin{enumerate}
    \item \textbf{Task 1a} involves predicting whether a given text is humorous. 
    \item \textbf{Task 1b} requires predicting the humor rating of a given humorous text.
    \item \textbf{Task 1c} incorporates humor subjectivity by posing a classification problem of predicting whether the underlying humor is controversial or not. 
\end{enumerate}
    \item \textbf{Task 2} is an offense detection task and is posed as a bounded regression problem. Given a text, we need to predict a mean score denoting the text's offensiveness 
    on a scale of $0$ to $5$, with $5$ being the most offensive.
\end{enumerate}

\begin{figure*}[t]
    \centering
    \begin{subfigure}[b]{0.48\textwidth}
        \includegraphics[width=\textwidth]{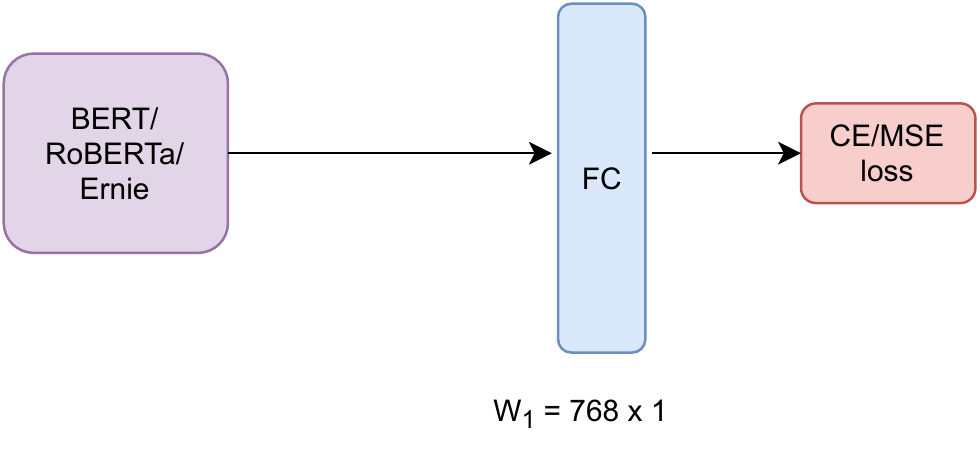}
        \caption{Single-task
        model}
        \label{fig:single_task_model}
    \end{subfigure}
    \hfill
    \begin{subfigure}[b]{0.48\textwidth}
        \includegraphics[width=\textwidth]{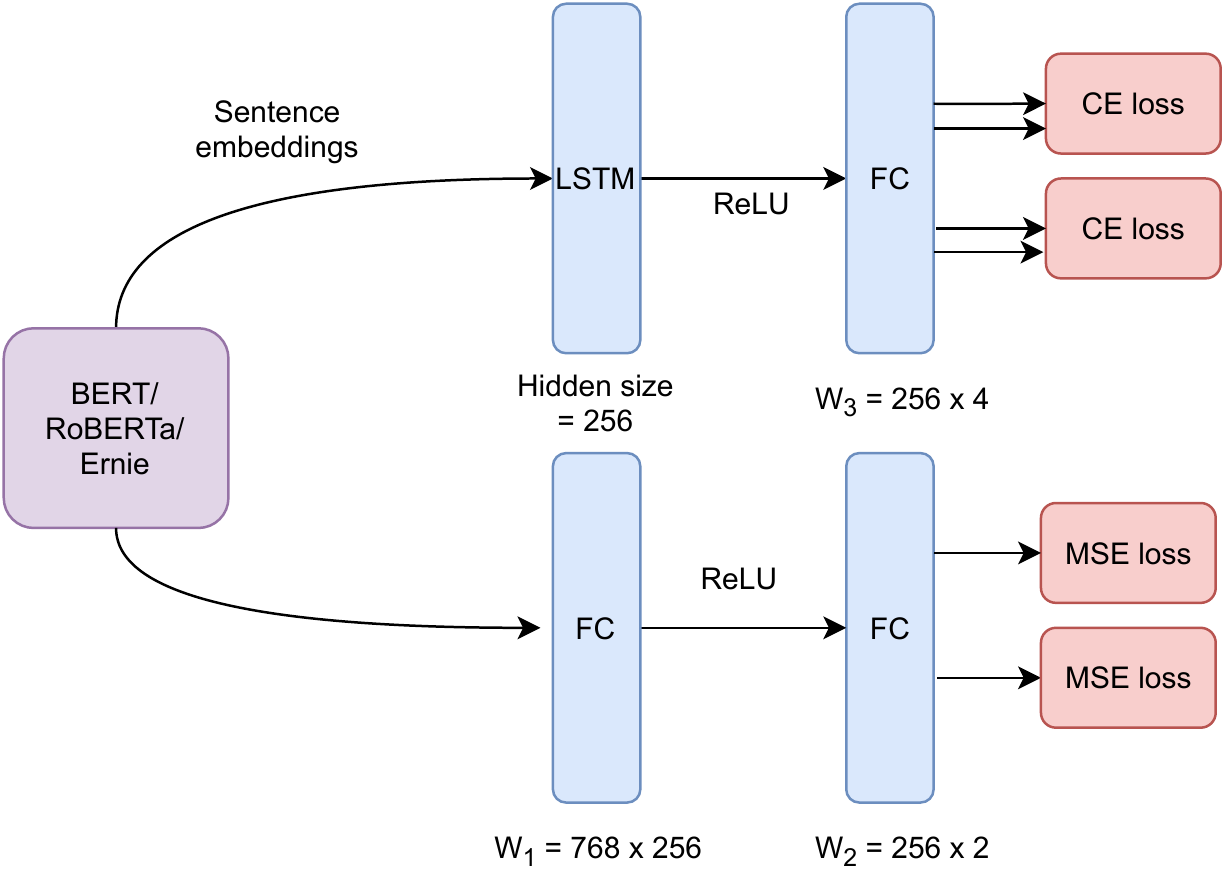}
        \caption{Multi-task model}
        \label{fig:mtl_model}
    \end{subfigure}
    \caption{Different Model architectures used for Humor/Offense detection/rating.}
    \label{fig:models}
\end{figure*}

\subsection{Related Works}

\textbf{Transfer Learning} ULMFiT~\citep{howard2018universal} used a novel neural network based method for transfer learning and achieved SOTA results on a small dataset. \citet{devlin2018bert} introduced BERT to learn latent representations in an unsupervised manner, which can then be finetuned on downstream tasks to achieve SOTA results. \citet{lan2019albert, liu2019roberta, sanh2019distilbert, sun2019ernie} have proposed several improvements to the BERT model. In this paper, we analyze the effects of using these different base models in the context of humor and offense detection.

\noindent \textbf{Humor \& Emotion Detection} \citet{humorDetection2019} first proposed the use of transformers ~\citep{vaswani2017attention} in humor detection and outperformed the state of the art models on multiple datasets.  \citet{ismailov2019humor, annamoradnejad2020colbert} extended the use of BERT models to humor classification. \citet{flescan-lovin-arseni-etal-2017-warteam} did humor classification by comparing and ranking tweets while \citet{docekal2020jokemeter} edit the tweet and rank the extent of humor for the edited tweet on a scale of 0 to 3 (most funny).  There has been extensive research in the area of text emotion prediction and generation (e.g., \citet{witon-etal-2018-disney,colombo-etal-2019-affect,goswamy-etal-2020-adapting,singh2021end}).  \citet{demszky2020goemotions} curated a large scale emotion detection dataset and achieved SOTA results by finetuning a BERT model. However, none of these works delve into humor analysis' subjectivity, which is a prime focus of this task.

\noindent \textbf{Sentiment and Pun Analysis} \citet{li2019exploiting, pmlr-v128-maltoudoglou20a} study BERT based models for sentiment analysis. \citet{ke2019sentilr} uses a combination of sentence embedding, POS tagging and word-level sentiment polarity scores for sentiment classification. \citet{zhou2020boating} uses contextualized and pronunciation embeddings for each word and pass these through a neural network to detect and localize pun in the sentence. However, none of these works focus on the subjectivity of the underlying sentiment and pun in the text.


\section{System Overview}
\label{sec:system}

\subsection{Data}
\label{subsec:data}
The challenge dataset comprises of a  \texttt{train} set (labeled 8000 texts) and a \texttt{public-dev} set (labeled 1000 texts). Each text input is labeled as $1/0$ if it is humorous or not and rated with the offensiveness score on a scale of $0$-$5$. If a text is classified as humorous, it is further annotated with humor rating and classified as controversial or not. For our single-task models (Section~\ref{subsec:st_models}), we train on the \texttt{train} + \texttt{public-dev} set after obtaining a suitable stopping epoch by training and validating on the \texttt{train} and \texttt{public-dev} respectively. For our multi-task models (Section~\ref{subsec:mtl_models}), we train on 8200 texts sampled randomly from \texttt{train} and \texttt{public-dev} sets and use remaining 800 text inputs for validation.




\subsection{Single Task Model}
\label{subsec:st_models}
As the tasks are evaluated independently, we have explored LLMs for each task/subtask independently and will be referring to them as single task models. Inspired by \citet{demszky2020goemotions}, for each task, we add a classification (for Task 1a, 1c) or a regression (for Task 1b, 2) head on top of the pretrained models like BERT, RoBERTa, ERNIE-2.0, DeBERTa and XLNet and train the model end-to-end (Figure \ref{fig:single_task_model}). This ensures that the model learns features solely related to the task, enhancing the performance. Also, as we only add a classification/regression head, the number of learnable parameters does not increase much. This helps us in finetuning the model on such a small dataset for a few number of epochs avoiding overfitting and resulting in better generalization.   

\subsection{Multi Task Learning}
\label{subsec:mtl_models}

\citet{10.1145/1390156.1390177} demonstrated that Multi-Task Learning (MTL) improves generalization performance across tasks in NLP. The different tasks though uncorrelated, share the same underlying data distribution. This can be of great help for tasks 1b and 1c where labeled instances are far less than for task 1a or 2. Exploiting the fact that all tasks share same data distribution, we propose to learn a model jointly on all the tasks. Specifically, we consider hard parameter sharing among differnet tasks and parameterize the base models using a neural network, followed by two heads for classification and regression tasks (Figure~\ref{fig:mtl_model}). Our base model includes LLMs like BERT, RoBERTa, and ERNIE. Contrary to the LSTM layer, which helps in learning features using all the token level embeddings, the Fully Connected (FC) layer focuses only on the embedding of [CLS] token. Hence, having these two branches allow the model to focus on different tasks using the same sentence embedding and helps in learning enhanced embeddings for task 1b and 1c with much lesser labeled dataset.

\subsection{Ensembles}
Mostly LLMs differ in their training procedure, and architecture. These big language model frameworks are trained on wide set of datasets for a variety of tasks. Though, they all have comparable performance, they may still capture different aspects of the input. We try to leverage such varied informative embeddings based predictions by combining multiple models trained with different basenet using following strategies:


\noindent \textbf{Jointly trained Model Embeddings:}
All the big language frameworks have shown huge performance improvement on multiple tasks owing to their highly informative latent input embeddings. We propose to learn an ensemble leveraging diverse aspects of the input captured by varied LLMs by concatenating their latent embeddings and mapping them to low dimensional space for task prediction. We use this method in learning ensembles of single task models explained in~\ref{subsec:st_models}. 

\begin{figure}[t]
    \centering
    \includegraphics[width=\linewidth]{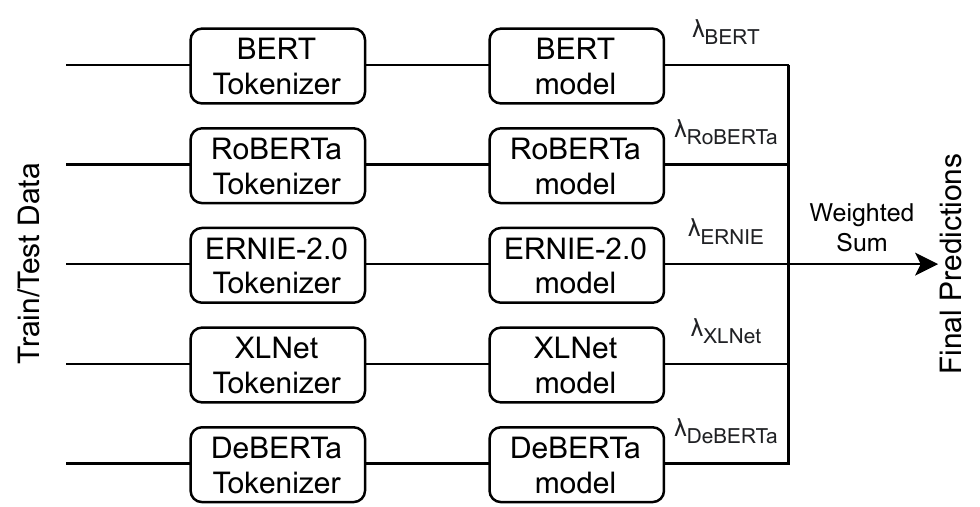}
    \caption{\textbf{Weighted-Average Ensembling:} The data is tokenized and then passed to the respective model. A weighted sum is done to obtain the final predictions. $\lambda_i$ represents the weight for model $i$.}
    \label{fig:ensemble}
\end{figure}


\begin{table*}[t]
    \begin{adjustbox}{width=2\columnwidth, center}
        \begin{tabular}{c c c c c c c c}
            \toprule
            \multirow{2}{*}{Model} &\multicolumn{2}{c}{Task1-a} & Task1-b & \multicolumn{2}{c}{Task1-c} & Task2 \\
            \cline{2-3}\cline{5-6} 
             & F-Score & Accuracy & RMSE & F-Score & Accuracy & RMSE \\
            \midrule
            STM (BERT) & - & - & 0.5841 & 0.5934 & 0.4829 & 0.4997 \\
            STM (RoBERTa) & 0.9523 & 0.9410 & 0.5929 & \textbf{0.6242} & 0.4536 & - \\
            STM (ERNIE-2.0) & 0.9541 & 0.9430 & 0.5546 & 0.4113 & 0.5252 & 0.4716 \\
            STM (XLNet) & - & - & 0.5656 & 0.5892 & 0.5171 & - \\
            STM (DeBERTa) & 0.9532 & 0.9420 & 0.5491 & - & - & - \\
            STM (Agg. Ensemble) & \textbf{0.9581} & \textbf{0.9480} & 0.5480 & 0.4520 & \textbf{0.6209} & 0.4750 \\
            MTM (BERT)      & 0.9374    & 0.9210  & 0.5794 & 0.5080 & 0.5496 & 0.5049 \\
            MTM (RoBERTa)   & 0.9477 & 0.9350  & 0.5873 & 0.5479 & 0.5170 & 0.5141 \\
            MTM (ERNIE-2.0)     & 0.9530 & 0.9420  & 0.5541 & 0.5389 & 0.5187 & 0.4961 \\
            STM + MTM (Agg. Ensemble)  & 0.9520 & 0.9400  & \textbf{0.5210} & 0.5321 & 0.5252 & \textbf{0.4520} \\
            \bottomrule
        \end{tabular}
    \end{adjustbox}
    \caption{Metrics on the test dataset for the major models on all the sub-tasks. MTM stands for Multi-Task Model, STM stands for Single Task Model, and Agg. Ensemble is Aggregation Based Ensembling without having to jointly train all the models together.}
    \label{tab:results}
\end{table*}

\noindent \textbf{Aggregation of Trained Model Predictions:} Joint-training though more informative and powerful, is a computationally intensive approach. Thus as an alternative, we use a weighted averaging of multiple pretrained models without compromising much on the performance.
\begin{enumerate}[topsep=0pt,itemsep=-1ex,partopsep=1ex,parsep=1ex]
    \item \textbf{Weighted Aggregate of Regression Outputs:}  For an ensemble of $k$ models trained using different LLMs as basenet, the aggregate output $\hat{y}$ is computed as $\hat{y} = \sum_{i = 1}^k \lambda_i \cdot \hat{y}_i \nonumber$ where $y_i$ and $\lambda_i$ represents the output and weight of the $i^{th}$ model respectively. The weights $\lambda_i$ are obtained through extensive grid search on the held out validation dataset or set to a $\frac{1}{k}$ when trained on the entire dataset without a validation set. The complete approach is shown in figure~\ref{fig:ensemble}.
    
    \item \textbf{Voting Based Classification:} This is one of the most popular approach of learning an ensemble and does not involve any hyperparameters or retraining of any of the constituent models. This involves training multiple models independently and using maximum among all the predictions as the final output. For a binary classification task, the final output $\hat{y}$ is by max-voting across the independent models.
\end{enumerate}


\begin{table*}[t]
    \begin{adjustbox}{width=2\columnwidth, center}
        \begin{tabular}{c c c c c c c}
            \toprule
            \multirow{2}{*}{Rank} & \multicolumn{2}{c}{Task1-a} & Task1-b & \multicolumn{2}{c}{Task1-c} & Task2 \\
            \cline{2-3}\cline{5-6}
             & F-Score & Accuracy & RMSE & F-Score & Accuracy & RMSE \\
            \midrule
            Rank-1 & 0.982 & 0.9854 & 0.4959 & 0.4943 & 0.6302 & 0.4120 \\
            Rank-2 & 0.975 & 0.9797 & 0.4977 & 0.4699 & 0.6279 & 0.4190 \\
            Rank-3 & 0.960 & 0.9676 & 0.5210  & 0.4699 & 0.6270  & 0.4230 \\
            Ours   & 0.948 (21) & 0.9581 (21) & 0.5210 (3) & 0.452 (9)  & 0.6209 (9) & 0.4607 (16)\\
            \bottomrule
        \end{tabular}
    \end{adjustbox}
    \caption{Comparison of our results with those on top of the leaderboard. (*) indicates our rank on the leaderboard in that task.}
    \label{tab:lboard_comparison}
\end{table*}

\section{Experimental Setup}
\label{sec:expt}

We used Pytorch~\citep{NEURIPS2019_9015} and HuggingFace~\citep{wolf2020huggingfaces} library for our models, and Google Colab GPUs for training and inference. We use ADAMW~\cite{loshchilov2019decoupled} and ADAM~\cite{kingma2017adam} optimizer with initial learning rate of $2e^{-5}$ for training single task  and multi task models respectively. For each of the models we follow a dedicated training pipeline described in subsequent sections.

\subsection{Data preprocessing}
We split the dataset into training and validation data as described in Section~\ref{subsec:data}. The sentences are annotated with a [CLS] token in the beginning and given as an input to the model. We performed additional experiments by removing stopwords but noticed a slight deterioration in the performance.

\subsection{Loss Functions}
Task 1a \& 1c are instances of binary classification problem and thus have been trained using cross-entropy loss. For predicting humor and offense rating i.e., Task 1b and 2, we have used mean squared error as the loss function.

\subsection{Training Details}
All the models are trained for $n$ epochs where $n$ is a hyper-parameter tuned on the validation set using early stopping criteria. For single task models, we split \texttt{train} data into training and validation set to learn the optimal value of $n$ and then train the model from scratch on \texttt{train} + \texttt{public-dev} set for $n$ epochs. In case of multi task models, all the tasks do not converge at the same rate. Thus, we train multi task models on randomly sampled $8200$ texts from \texttt{train} + \texttt{public-dev} dataset and validate on the remaining 800 texts. We use early stopping criteria on validation dataset independently for each task.

\section{Results}
\label{sec:results}

We have trained multiple single task and multi task models using basenet LLMs like  BERT, DistilBERT, RoBERTa, XLNet, Albert \citep{lan2019albert}, Electra \citep{clark2020electra}, DeBERTa, and ERNIE-2.0. We also learned ensembles of single task models by either training a classification/regression head on concatenated input embeddings or using weighted aggregate of the models' predictions. Apart from this, we also explored voting based ensemble of multi-task models. All our models perform comparably on all tasks and the major models are reported in Table~\ref{tab:results}. We also compare our best model performance with the top $3$ submissions on the leaderboard and report it in Table~\ref{tab:lboard_comparison}.
\section{Analysis}
\label{sec:analysis}
\subsection{Data Augmentation}

One recurring issue across all our trained models is the high susceptibility to overfitting. Data Augmentation is a widely accepted solution to reduce overfitting by generating slight variants of the given dataset and is extremely useful for a smaller dataset.

One such approach is Masked Language Modelling (MLM), used to perform context-specific data augmentation~\citep{ma2019nlpaug} and has been used in training LLMs. However, following this data augmentation during training has consistently degraded the performance of our models. We hypothesize that this is due to the mismatch between the contextual meaning and the associated humor/offense. MLM-based augmentation strategies, with models pre-trained to preserve the sentence's meaning, fail to capture the associated humor/offense.

Often the selection of words in a sentence is responsible for its humor/offensive rating. Replacing such words by their synonyms can change the humor/offense rating substantially. Hence, using such a data augmentation approach during training will inject heavy noise in the ground truth resulting in deteriorated performance.


\subsection{Correlation across Tasks}

Contrary to our belief, we fail to ascertain any direct relationship between the humor controversy and the offense rating prediction task. We compute the mean offense rating for the texts labeled as controversial and for texts marked as non-controversial. The computed mean values are too close to each other to demonstrate any direct correlation conclusively.

\subsection{Dataset Size}
In literature, finetuning LLMs on small size task specific dataset has shown remarkable task performance. However, our single dedicated task models could not perform better than our multi-task model for Task 1b. We attribute this to relatively small size of supervised dataset available for Task 1b incomparison to other tasks. In our multi task models, though we have lesser labeled text for Task 1b, our sentence embeddings are still updated using the complete available dataset. Thus, our multi task model learns underlying distribution better than single task model owing to join learning and shared parameters for task 1b and 2. We believe that this is the main reason for the enhanced performance of our model on Task 1b which has lesser supervised data available in comparison to Task 1a or 2.


\section{Conclusion}
\label{sec:conclude}

We have presented several experiments using large language models like BERT, XLNet, etc., and their ensembles for humor and offense detection and rating. We also discuss some of the underlying challenges due to the subjective nature of humor and offense detection task. Using these, we explain why standard training practices used to prevent overfitting, like data augmentation, do not work in this context. Our experiments suggest that even though these models can reasonably capture humor and offense, they are still far from understanding every intricacy arising out of subjectivity. To tackle some of the problems highlighted in this paper, a compelling direction would be online data augmentation by alternating between training the embeddings and generating new texts to preserve the humor/offensiveness. Additionally, pretraining these models on datasets annotated by diverse annotators to capture a more comprehensive world knowledge should further help in generalization.

\bibliography{acl2020}
\bibliographystyle{acl_natbib}


\end{document}